\documentclass[runningheads]{llncs}
\usepackage{graphicx}
\usepackage{color}
\usepackage{times}
\usepackage{epsfig}
\usepackage{amsmath}
\usepackage{amssymb}
\usepackage{multirow}
\usepackage{bm}
\begin{document}
\title{Synthetic Sample Selection via Reinforcement Learning}
%
%
\author{Jiarong Ye\inst{1}\thanks{These authors contributed equally to this work.}, Yuan Xue\inst{1}\textsuperscript{*}, L. Rodney Long\inst{2},  Sameer Antani\inst{2}, Zhiyun Xue\inst{2}, \\ Keith C. Cheng\inst{3}, Xiaolei Huang\inst{1}}
\institute{College of Information Sciences and Technology, The Pennsylvania State University, University Park, PA, USA
\and National Library of Medicine, National Institutes of Health, Bethesda, MD, USA
\and College of Medicine, The Pennsylvania State University, Hershey, PA, USA}
\authorrunning{Jiarong Ye~\textit{et al.}}

%
\maketitle              

\begin{abstract}
Synthesizing realistic medical images provides a feasible solution to the shortage of training data in deep learning based medical image recognition systems. However, the quality control of synthetic images for data augmentation purposes is under-investigated, and some of the generated images are not realistic and may contain misleading features that distort data distribution when mixed with real images. Thus, the effectiveness of those synthetic images in medical image recognition systems cannot be guaranteed when they are being added randomly without quality assurance. In this work, we propose a reinforcement learning (RL) based synthetic sample selection method that learns to choose synthetic images containing reliable and informative features. A transformer based controller is trained via proximal policy optimization (PPO) using the validation classification accuracy as the reward. The selected images are mixed with the original training data for improved training of image recognition systems. To validate our method, we take the pathology image recognition as an example and conduct extensive experiments on two histopathology image datasets. In experiments on a cervical dataset and a lymph node dataset, the image classification performance is improved by $8.1\%$ and $2.3\%$, respectively, when utilizing high-quality synthetic images selected by our RL framework. Our proposed synthetic sample selection method is general and has great potential to boost the performance of various medical image recognition systems given limited annotation.


\end{abstract}

\section{Introduction}
The success of deep learning in vision tasks heavily relies on large scale datasets with high quality annotations~\cite{deng2009imagenet,lin2014microsoft}. However, in the domain of medical images, large scale datasets, especially with expert annotations, are often unavailable due to the high cost of annotation and privacy concerns. To mitigate the data insufficiency in medical image tasks, one feasible solution is to manually augment the original dataset. However, random augmentation~\cite{wang2017effectiveness} including random spatial and intensity augmentation are difficult to generalize and sometimes generate unrealistic images. Recently, pioneering works~\cite{bowles2018gan,chaitanya2019semi,frid2018gan,gupta2019generative,zhao2019data} have been done using synthetic data generated by Generative Adversarial Network (GAN) models to augment medical training datasets with limited annotation. While achieving promising results, previous works mainly focus on advancing the image synthesis models and often neglect the quality assessment of generated images in terms of their effectiveness in downstream tasks such as improved image recognition. 
Unlike real images, synthetic images are with unequal qualities and many of them may contain features that skew the data distribution.
Thus, directly expanding the original dataset with synthetic images cannot guarantee performance improvement on various downstream tasks such as image recognition and classification. 

In this work, we aim at complementing the medical image synthesis with synthetic sample selection, \textit{i.e.}, how to select high quality synthetic samples to improve medical image recognition systems. The first step of our method is synthesizing images for selection. For small training sets, advanced images synthesis models include conditional GAN (cGAN)~\cite{mirza2014conditional} which generates fake images from class label and noise vector, and cycleGAN~\cite{zhu2017unpaired} which translates existing images into new synthetic images. We focus on the cGAN setting and design a new cGAN model for histopathology image synthesis. For synthetic sample selection, a recent work~\cite{xue2019synthetic} has attempted to use distance to class centroids in the feature space as a criterion to select synthetic images; however, their approach is not learning based and lacks generality. 

Considering that the selection of synthetic images can be modeled as a binary decision making problem, the decision-making criteria can be either handcrafted or learned. Compared to using a handcrafted metric such as that in~\cite{xue2019synthetic}, an agent trained with a learning algorithm can
automatically make decision based on a more comprehensive pool of learned features and classification performance gains, thus is more likely to achieve superior performances in downstream tasks. Intuitively, such selection mechanism can be framed as a model-free, policy-based reinforcement learning (RL) process, in which a controller determines the actions applied to the candidate synthetic images and keeps updating based on the reward of
downstream task performance until convergence. Therefore, we propose a reinforcement learning based selection model to automatically choose the most representative images with highest quality. The backbone of our selection model is a combination of ResNet34~\cite{he2016deep} model for feature extraction and a transformer~\cite{vaswani2017attention} model for sample selection. We use the validation classification accuracy as the reward of RL training and use the state-of-the-art proximal policy optimization (PPO)~\cite{schulman2017proximal} to update the model weights.

To the best of our knowledge, our work is the first to systematically investigate the synthetic sample selection problem for medical images. Our proposed method is validated on two histopathology datasets~\cite{xue2019synthetic,veeling2018rotation} with limited number of annotations. Experimental results show that our method substantially outperform all previous methods including traditional augmentation, direct synthetic augmentation with GAN-generated images, and selective augmentation based on feature distances. With RL as the learning engine, our proposed synthetic sample selection method provides a principled approach to a critical missing link in medical image synthetic augmentation.
Moreover, comparison between selected and discarded samples can help us gain deeper understanding about which types of feature encode semantically-relevant information. 

\begin{figure}[ht]
\begin{center}
  \includegraphics[width=0.90\linewidth]{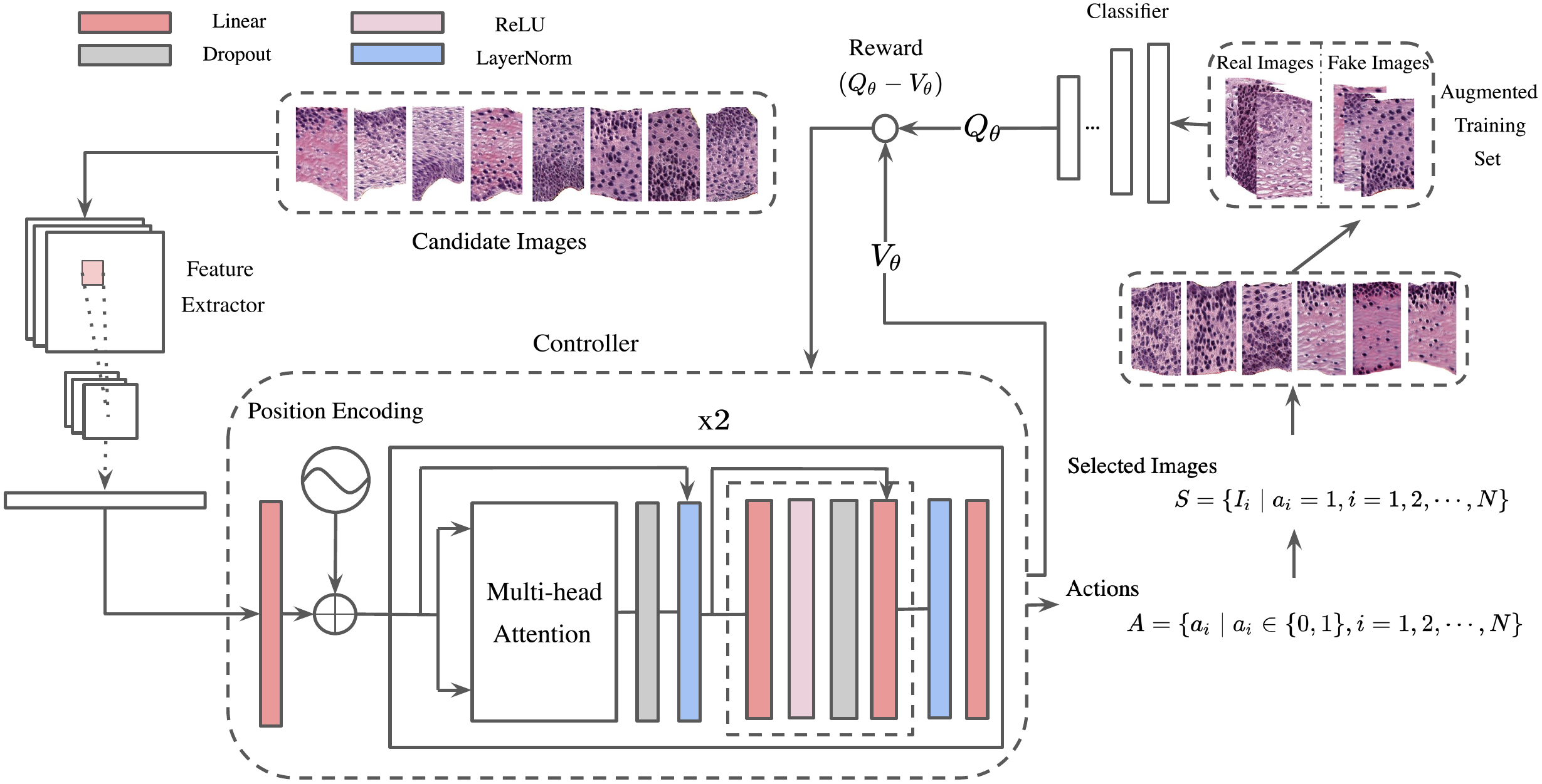}
\end{center}
  \caption{The architecture of our proposed selective synthetic augmentation framework based on Reinforcement Learning.}
\label{fig:architecture}
\end{figure}

\section{Methodology}

The detailed architecture of our proposed framework is shown in Figure~\ref{fig:architecture}. The candidate pool of synthetic images is generated with HistoGAN designed for synthesizing histopathology patches (details are provided in supplementary materials). The controller which is a transformer model~\cite{vaswani2017attention} plays a vital role of deciding whether to select a synthetic sample or not. In each training iteration, it takes the feature vectors extracted from a ResNet34~\cite{he2016deep} model trained on the original training images as the input, then outputs the binary action (select or discard) for each candidate synthetic image. After sample selection, 
we train the classifier on the expanded dataset and take the maximum validation accuracy of last five epochs~\cite{zoph2016neural} as the the reward to update the policy.
To improve the stability of policy update and avoid severe fluctuation caused by parameter change, we adopt Proximal Policy Optimization (PPO)~\cite{schulman2017proximal} as the policy gradient method in our synthetic sample selection model. 
Next, we will introduce the detailed design of the controller and policy gradient method, which are two main components in our proposed framework. 

\subsection{Controller}

The rationale behind the choice of the controller is based on the feature dependencies among candidate images. We hypothesize that the order of augmentation is not entirely independent since the late adds are subject to a constraint to differentiate with the early adds to assure diversity of the entire augmented training set. 
In order to address the potential relation existing between images augmented while avoiding an strong sequential assumption, we leverage the self-attention mechanism by adopting the transformer~\cite{vaswani2017attention} model as the controller. In the transformer architecture, all recurrent structure are eschewed, thus the feature vectors need to be combined with their embedded positions based on sinusoidal functions as the input to the encoder layer of transformer. The main component of the encoder inside transformer is the multi-head attention block that consists of $n$ self-attention layers, where $n$ refers to the number of heads.
In each self-attention layer, input features are projected to three feature spaces as query, key and value by multiplying learnable weight matrices. And the attention map is obtained by the following equation:
\begin{small}
\begin{equation}
\text{Attention}(Q, K, V) = \text{softmax} \left ( \frac{QK^T}{\sqrt{d_k}}V \right )\enspace .
\label{Eq:self-attn}
\end{equation}
\end{small}
Each head represents a different projected feature space for the input, with the same input embedding multiplying different weight matrix, then concatenated at the end to generate the final attention map as:
\begin{small}
\begin{equation}
\text{MultiHead}(Q, K, V) = 
[\text{head}_1;\text{head}_2]
W^O\enspace,
\label{Eq:multi-head}
\end{equation}
\end{small}
$\text{where head}_i = \text{Attention}(QW_i^Q, KW_i^K, VW_i^V)$. The learnable weight matrices are denoted as: $W^Q \in \mathbb{R}^{d_{\text{input}}\times d_k}$, $W^K \in \mathbb{R}^{d_{\text{input}}\times d_k}$, $W^V \in \mathbb{R}^{d_{\text{input}}\times d_v}$, $W^O \in \mathbb{R}^{hd_{v}\times d_{\text{input}}}$, here $h$ represents the number of heads. Then after applying the attention map, the context vector is fed into the feed forward layer as follows:
\begin{small}
\begin{equation}
    F(x) = \max(0, xW_1 + b_1)W_2 + b_2 \enspace.\label{Eq:feed-forward}
\end{equation}
\end{small}

Skip connections are adopted in the process to combine learned features from both the higher and lower abstract levels.
Considering the task of the controller is to output a binary action for each input feature vector, as shown in Fig.~\ref{fig:architecture}, the decoder for transformer is a linear layer used as the policy network. All in all, applying the transformer as the controller for our reinforcement learning based image selection framework is advantageous for its self-attention mechanism to capture the dependencies among input feature vectors. We perform comprehensive ablation study over the choice of controller and results are presented in Section~\ref{sec:results}.

\subsection{Policy Gradient Method}\label{section:pg}

An effective while efficient policy gradient method is crucial for the entire reinforcement learning process to optimally leverage the reward as the feedback to the controller. Proximal Policy Optimization (PPO)~\cite{schulman2017proximal} has become a popular choice of policy gradient algorithm, due to its computational efficiency and satisfactory performance compared with previous algorithms such as TRPO~\cite{schulman2015trust}. In our framework we use the PPO to stabilize the RL training process. To reduce complexity while maintaining comparable performance, in the algorithm of PPO, the KL convergence constraint enforced in TRPO on the size of policy update is replaced with clipped probability ratio between the current and the previous policy within a small interval around 1. At time step $t$, let $A_{\theta}$ be the advantage function, the objective function is as follows:
\begin{small}
\begin{equation}
    \mathcal{L(\theta)} = \mathbb{E} \left [ \min(\gamma_{\theta}(t) A_{\theta}(s_t, a_t), \text{clip} (\gamma_{\theta}(t), 1-\epsilon, 1+\epsilon) A_{\theta}(s_t, a_t)) \right ]\enspace,
    \label{Eq:ppo_obj}
\end{equation}
\end{small}
where $A_{\theta}(s_t, a_t) = Q_{\theta}(s_t, a_t) - V_{\theta}(s_t, a_t)$. As part of the transformer output, $V_{\theta}(s_t, a_t)$ is a learned state-value taken off as the baseline from the q-value to lower the variation of the rewards along the training process. $\pi$ refers to the probability of actions. $Q_{\theta} (s_t, a_t)$ is the q-value at time $t$ defined as the smooth version of the max validation accuracy among the last 5 epochs in the classification task. Since our target tasks are trained on limited number of data, in order to get a robust estimation of reward changing pattern, we apply the Exponential Moving Average (EMA) \cite{hunter1986exponentially} algorithm to smooth the curve of original reward. With smoothing, the final reward at time $t$ is: 
\begin{small}

\begin{equation}
    \hat{Q}_{\theta}(s_t, a_t) = \left\{\begin{matrix} \text{Q}_{\theta}(s_t, a_t),& \enspace t=1
\\ \alpha \hat{Q}_{\theta}(s_{t-1}, a_{t-1}) + (1-\alpha)\text{Q}_{\theta}(s_t, a_t),& \enspace t>1  \label{Eq:smoothReward}
\end{matrix}\right. \enspace.
\end{equation}
\end{small}
Based on the idea of importance sampling, the weight assigned to the current policy also depends on older policies. The probability ratio between previous and current policies $\gamma_{\theta}(t)$ is defined by :
\begin{small}
\begin{equation}
    \gamma_{\theta}(t) = \frac{\pi_{\theta}(a_t \mid s_{t})}{\pi_{\theta} (a_{t-1} \mid s_{t-1})} \enspace,
    \label{Eq:ppo_ratio}
\end{equation}
\end{small}
where $a_t \in \mathbb{R}^{N \times 2}$, $N$ refers to the number of synthetic images in the candidate pool. If at time step $t$, 
$a_{i}(t)=0, i \in \{1,2, \cdots N\}$, then the candidate $i$ is discarded, otherwise it is added to the original training set for further steps. We also compare PPO with the classic REINFORCE~\cite{williams1992simple} algorithm in experiments.

\section{Experiments}

\begin{figure}[t]
\begin{center}
  \includegraphics[width=0.95\linewidth]{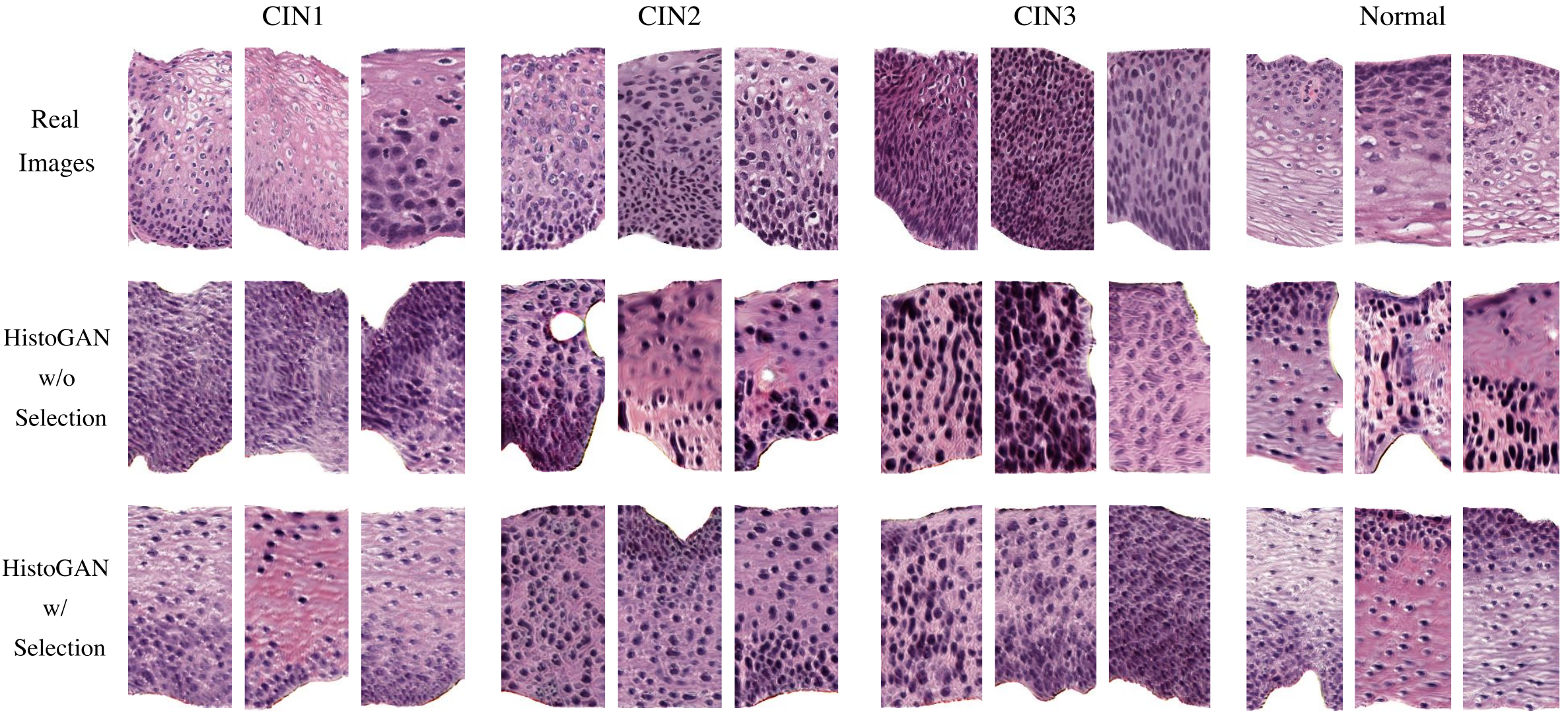}
\end{center}
  \caption{Examples of real images and synthetic images generated by HistoGAN trained on cervical histopathology dataset and selected with our RL-based framework.  Zoom in for better view.}
\label{fig:histology}
\end{figure}

\begin{figure}[t]
\begin{center}
  \includegraphics[width=0.95\linewidth]{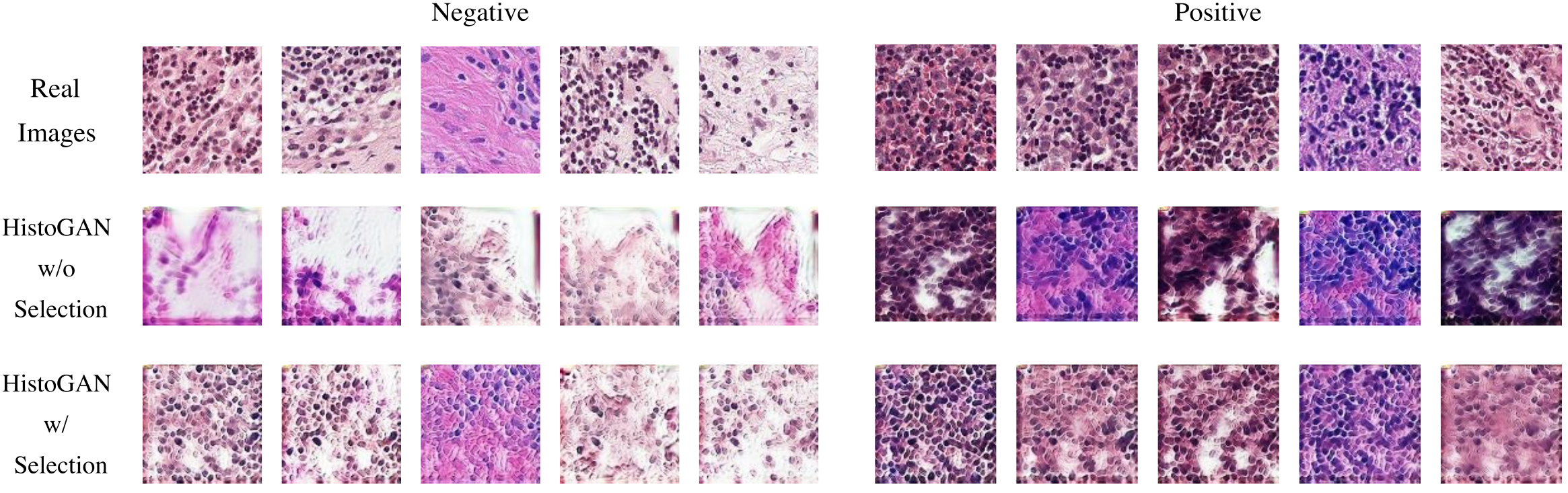}
\end{center}
  \caption{Examples of real images and synthetic images generated by our HistoGAN model trained on 3\% of PCam dataset and selected with our framework.}
\label{fig:pcam}
\end{figure}


We conduct comprehensive experiments on two histopathology datasets. The first dataset is a cervical histopathology dataset where all images are annotated by the same pathologist. The data processing follows~\cite{xue2019synthetic}, and results in patches with a unified size of $256 \times 128$ pixels. Compared with the dataset used in~\cite{xue2019synthetic}, we include more data for more comprehensive experiments. In total, there are $1,284$ Normal, $410$ CIN1, $481$ CIN2, $472$ CIN3 patches. Examples of the images can be found in Fig.~\ref{fig:histology}. 
We randomly split the dataset, by patients, into training, validation, and testing sets, with ratio 7:1:2 and keep the ratio of image classes almost the same among different sets. All evaluations and comparisons reported in this section are carried out on the test set.

To further prove the generality of our proposed method, we also conduct experiments on the public PatchCamelyon (PCam) benchmark~\cite{veeling2018rotation}, which consists of patches extracted from histopathologic scans of lymph node sections with unified size of $96 \times 96$ pixels.
To simulate the scenario of limited training data, we use the randomly selected $3\%$ of the training set in PCam, which results $3,931$ negative and $2,757$ positive patches, to train the HistoGAN model and the baseline classifier. All trained models are evaluated on the full testing set. Example results are illustrated in Fig.~\ref{fig:pcam}.

\subsection{Implementation Details}

As illustrated in Fig.~\ref{fig:architecture}, the candidate pool contains $1,024$ HistoGAN-generated images for each class (CIN1, CIN2, CIN3, NORMAL) of the cervical histopathology dataset, and $2,048$ images for each class (Negative, Positive) of the PCam dataset. These candidates are fed into a ResNet34 model pretrained on the original dataset and the feature vectors of the fully-connected layer are extracted as input for the controller. The entire set of input is first sorted by the cosine distances to the corresponding centroids and divided into 8 batches, with each containing $128$ images from each class for histopathology dataset, and $256$ images from each class for PCam dataset.  
The controller consists of 2 encoder layers, in which the multi-head attention block contains 2 heads. It outputs a binary action vector $A_{\theta}$ for the further selection of the augmented training set, as well as a value $V_{\theta}$ used in the calculation of reward for the policy gradient method. The classifier is also a ResNet34 model of the same structure as the feature extractor. 
Similar to~\cite{zoph2016neural}, we adopt the max accuracy obtained from testing on the validation set of the last 5 epochs in the classification task. 
To further stabilize the training reward, we use the EMA-smoothed max validation accuracy as the reward with $\alpha=0.8$. In the policy gradient algorithm PPO, the policy function $\pi$ is obtained from the softmax layer of the ResNet34 model, after the policy network, \textit{i.e.}, a linear layer with output dimension set to the number of classes in the corresponding dataset. $\epsilon$ in Eq.~\ref{Eq:ppo_obj} is set to $0.2$, providing the upper and lower bound for the ratio of policy functions at current time step $t$ and previous time step $t-1$. In all synthetic augmentation methods, the augmentation ratio is set to $0.5$ to be consistent with~\cite{xue2019synthetic}. In our RL based method, the number of selected images are determined by the controller. The learning rate for the reinforcement learning framework training is $2.5e-04$, and for the attention mechanism in the ablation study in Table \ref{tb:cervical2} is $1e-04$.

\begin{table*}[ht]

\begin{center}
\resizebox{0.75\linewidth}{!}{
\begin{tabular}{|l|c|c|c|c|}
\hline
{} &     Accuracy &              AUC &      Sensitivity &      Specificity \\
\hline
Baseline Model~\cite{he2016deep}  &  
.754 $\pm$ .012 &  
.836 $\pm$ .008 &  
.589 $\pm$ .017 &  
.892 $\pm$ .005 \\

\hline
\enspace + Traditional Augmentation &  
.766 $\pm$ .013 &  
.844 $\pm$ .009 &  
.623 $\pm$ .029 &  
.891 $\pm$ .006 \\

\hline
\enspace + GAN Augmentation &
.787 $\pm$ .005 &  
.858 $\pm$ .003 &  
.690 $\pm$ .014 &  
.909 $\pm$ .003 \\

\hline
\enspace + Metric Learning & &  &  &  \\
\enspace \enspace  (Triplet Loss)~\cite{schroff2015facenet} 
&
\multirow{-1.7}*{.798 $\pm$ .016} &  
\multirow{-1.7}*{.865 $\pm$ .010} &  
\multirow{-1.7}*{.678 $\pm$ .048} &  
\multirow{-1.7}*{.909 $\pm$ .013} \\

\hline
\enspace + Selective Augmentation & &  &  &  \\
\enspace \enspace (Centroid Distance)~\cite{xue2019synthetic}$^{*}$ &
\multirow{-1.7}*{.808 $\pm$ .005} &  
\multirow{-1.7}*{.872 $\pm$ .004} &  
\multirow{-1.7}*{.639 $\pm$ .015} &  
\multirow{-1.7}*{.912 $\pm$ .006} \\

\hline
\enspace + Selective Augmentation & &  &  &  \\
\enspace \enspace (Transformer-PPO, Ours)&  
\multirow{-1.7}*{\textbf{.835 $\pm$ .007}} & 
\multirow{-1.7}*{\textbf{.890 $\pm$ .005}} &  
\multirow{-1.7}*{\textbf{.747 $\pm$ .013}} &  
\multirow{-1.7}*{\textbf{.936 $\pm$ .003}} \\

\hline
\end{tabular}
}
\end{center}
\caption{Classification results of baseline and augmentation models on the cervical dataset. We reimplemented~\cite{xue2019synthetic} and a metric learning model with triplet loss~\cite{schroff2015facenet} using the same pool of synthetic images generated by HistoGAN for fair comparison.}
\label{tb:cervical1}
\end{table*}

\subsection{Result Analysis}\label{sec:results}

In Fig.~\ref{fig:histology} and Fig.~\ref{fig:pcam}, we show qualitative results of synthetic images on cervical and lymph node datasets generated by HistoGAN, and images selected by our proposed synthetic sample selection model. From visual results, images selected by our method clearly contain more realistic features than images before selection. With HistoGAN generated images as candidates, we compare our RL based synthetic sample selection with traditional augmentation method~\cite{wang2017effectiveness} and other synthetic augmentation methods in Table~\ref{tb:cervical1} and Table~\ref{tb:cervical2}. The Traditional Augmentation includes horizontal flipping and color jittering of original training data; GAN Augmentation refers to randomly adding HistoGAN generated images to the original training set; In the Metric Learning method, we adopt the Triplet Loss~\cite{schroff2015facenet} which minimizes the intra-class differences and maximizes the inter-class differences between samples. We also compare with the current state-of-the-art sample selection method~\cite{xue2019synthetic} which ranks samples based on distance to class centroids in feature space. We report quantitative evaluation scores of all methods using the accuracy, area under the ROC curve (AUC), sensitivity and specificity. All models are run for $5$ rounds with random initialization for fair comparison. The mean and standard deviation results of the $5$ runs are reported. For fair comparison, we use the same backbone ResNet34 classifier with same hyperparameters setting in all experiments and use the same candidate pool generated by HistoGAN for sample selection to ensure that differences only come from the augmented training set.

\begin{figure}[ht]
\begin{center}
  \includegraphics[width=0.95\linewidth]{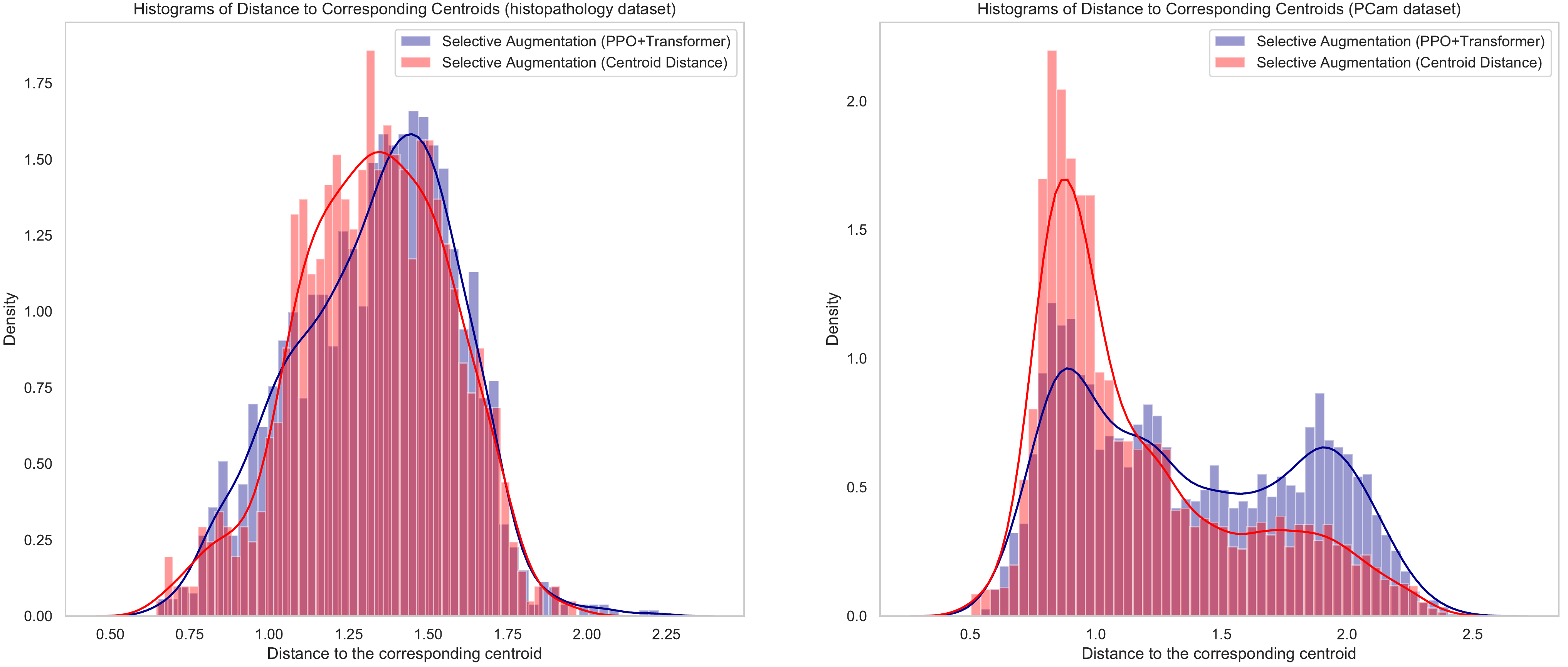}
\end{center}
  \caption{Comparison between histograms of the distances to the corresponding centroid of each class in the cervical histopathology dataset and PCam dataset.
  }
\label{fig:histogram}
\end{figure}

\begin{table*}[ht]

\begin{center}
\resizebox{0.75\linewidth}{!}{
\begin{tabular}{|l|c|c|c|c|}
\hline
{} &     Accuracy &              AUC &      Sensitivity &      Specificity \\
\hline
Baseline Model~\cite{he2016deep}  &  
.853 $\pm$ .003 &
.902 $\pm$ .002 &
.815 $\pm$ .008 &
.877 $\pm$ .009 \\

\hline
\enspace + Traditional Augmentation &  
.860 $\pm$ .005 &
.907 $\pm$ .003 &
.823 $\pm$ .015 &
.885 $\pm$ .017 \\

\hline
\enspace + GAN Augmentation &  
.859 $\pm$ .001 &
.906 $\pm$ .001 &
.822 $\pm$ .014 &
.884 $\pm$ .011 \\

\hline
\enspace + Metric Learning & &  &  &  \\
\enspace  \enspace  (Triplet Loss)~\cite{schroff2015facenet} &  
\multirow{-1.7}*{.864 $\pm$ .004} &
\multirow{-1.7}*{.910 $\pm$ .003} &
\multirow{-1.7}*{.830 $\pm$ .012} &
\multirow{-1.7}*{.887 $\pm$ .008} \\

\hline
\enspace + Selective Augmentation & &  &  &  \\
\enspace  \enspace (Centroid Distance) ~\cite{xue2019synthetic}$^{*}$ &
\multirow{-1.7}*{.868 $\pm$ .002} &
\multirow{-1.7}*{.912 $\pm$ .002} &
\multirow{-1.7}*{.835 $\pm$ .010} &
\multirow{-1.7}*{.890 $\pm$ .006} \\

\hline
\enspace + Selective Augmentation & &  &  &  \\
\enspace  \enspace (Transformer-PPO, Ours)&  
\multirow{-1.7}*{\textbf{.876 $\pm$ .001}} &
\multirow{-1.7}*{\textbf{.917 $\pm$ .001}} &
\multirow{-1.7}*{\textbf{.846 $\pm$ .010}} & 
\multirow{-1.7}*{\textbf{.895 $\pm$ .005}} \\

\hline
\end{tabular}
}
\end{center}
\caption{Classification results of baseline and augmentation models on the PCam dataset.} 
\label{tb:pcam}
\end{table*}

From Table~\ref{tb:cervical1} and Table~\ref{tb:pcam}, one can observe that compared with previous augmentation and sample selection methods, our proposed selective augmentation delivers superior result on both datasets using the same candidate synthetic image pool. While  achieving markedly better performances than the second best method~\cite{xue2019synthetic} in all metrics, we further analyse how images selected by our learning based method differ from those selected by the previous handcrafted sample selection method~\cite{xue2019synthetic}. We compare the distribution of selected images over the feature distance to corresponding class centroids. As shown in Fig.~\ref{fig:histogram}, both methods discard samples with too small or too large centroid distance as they are either too similar to original training data, or are outliers.
Especially on PCam dataset, \cite{xue2019synthetic} tends to select samples with relatively lower centroid distances. Meanwhile, our controller learns to select samples distributed evenly on different centroid distances and achieves promising results. We hypothesizes that a learning based method selects samples in a more general way than handcrafted method, and the histogram analysis further proves that handcrafted method based on a single metric may not be able to find an optimal pattern for sample selection. 

To validate our choice of the Transformer and PPO, we perform ablation study on the cervical dataset and report results in Table~\ref{tb:cervical2}. For the controller, we experiment with GRU~\cite{chung2014empirical} and GRU with attention~\cite{yang2016hierarchical} (GRU-Attn); For the policy gradient algorithm, we compare PPO with the classic REINFORCE~\cite{williams1992simple}. Compared with other controller and policy gradient algorithms, our full model with transformer and PPO achieves best performances in all metrics, which justifies our choices.

\begin{table*}[ht]
\begin{center}
\resizebox{0.75\linewidth}{!}{
\begin{tabular}{|l|c|c|c|c|}
\hline
{} &     Accuracy &              AUC &      Sensitivity &      Specificity \\
\hline
Selective Augmentation & &  &  &  \\
\enspace (GRU-REINFORCE) &  
\multirow{-1.7}*{.789 $\pm$ .011} &
\multirow{-1.7}*{.859 $\pm$ .007}  &
\multirow{-1.7}*{.687 $\pm$ .014} &
\multirow{-1.7}*{.908 $\pm$ .005} \\

\hline
Selective Augmentation & &  &  &  \\
\enspace (GRU-Attn-REINFORCE) &
\multirow{-1.7}*{.804 $\pm$ .019} &
\multirow{-1.7}*{.869 $\pm$ .012} &
\multirow{-1.7}*{.674 $\pm$ .039} &
\multirow{-1.7}*{.914 $\pm$ .010} \\

\hline
Selective Augmentation & &  &  &  \\
\enspace (Transformer-REINFORCE) &  
\multirow{-1.7}*{.812 $\pm$ .008} &
\multirow{-1.7}*{.875 $\pm$ .005} &
\multirow{-1.7}*{.724 $\pm$ .022} & 
\multirow{-1.7}*{.920 $\pm$ .006} \\

\hline
Selective Augmentation & &  &  &  \\
\enspace (GRU-PPO) &  
\multirow{-1.7}*{.792 $\pm$ .017} &
\multirow{-1.7}*{.862 $\pm$ .012} &
\multirow{-1.7}*{.701 $\pm$ .039} & 
\multirow{-1.7}*{.912 $\pm$ .010} \\

\hline
Selective Augmentation & &  &  &  \\
\enspace (GRU-Attn-PPO) &
\multirow{-1.7}*{.811 $\pm$ .014} &
\multirow{-1.7}*{.874 $\pm$ .010} & 
\multirow{-1.7}*{.751 $\pm$ .034} &
\multirow{-1.7}*{.919 $\pm$ .007} \\

\hline
Selective Augmentation & &  &  &  \\
\enspace (Transformer-PPO) &  
\multirow{-1.7}*{\textbf{.835 $\pm$ .007}} &
\multirow{-1.7}*{\textbf{.890 $\pm$ .005}} &
\multirow{-1.7}*{\textbf{.747 $\pm$ .013}} &
\multirow{-1.7}*{\textbf{.936 $\pm$ .003}} \\

\hline
\end{tabular}
}
\end{center}
\caption{Ablation study of our proposed reinforcement learning framework for synthetic images selection on the cervical histopathology dataset.}
\label{tb:cervical2}
\end{table*}

\section{Conclusions}

In this paper, we propose a reinforcement learning based synthetic sample selection method. Compared with previous methods using handcrafted selection metrics, our proposed method achieves state-of-the-art results on two histopathology datasets. In future works, we expect to extend our method to more medical image recognition tasks where annotations are limited. We also plan to investigate the usage of our RL based method for sample selection toward other purposes such as active learning for annotation. 

\section{Acknowledgements}
This work was supported in part by the Intramural Research Program of the National Library of Medicine and the National Institutes of Health. We gratefully acknowledge the help with expert annotations from Dr. Rosemary Zuna, M.D., of the University of Oklahoma Health Sciences Center, and the work of Dr. Joe Stanley of Missouri University of Science and Technology that made the histopathology data collection possible.
%
%
%
%
\bibliographystyle{splncs04}
\bibliography{ref}

\end{document}